# Generating Synthetic Photogrammetric Data for Training Deep Learning based 3D Point Cloud Segmentation Models

**Meida Chen, Andrew Feng, Kyle McCullough, Pratusha Bhuvana Prasad,**
***USC Institute for Creative Technologies***
**Los Angeles, California**
**{mechen, feng, McCullough, bprasad}@ict.usc.edu**

**Ryan McAlinden**
***Synthetic Training Environment Cross Functional Team***
**Los Angeles, California**
**ryan.e.mcalinden.civ@mail.mil**

**Lucio Soibelman**
***USC Department of Civil and Environmental Engineering***
**Los Angeles, California**
**soibelman@usc.edu**

**ABSTRACT**

At I/ITSEC 2019, the authors presented a fully-automated workflow to segment 3D photogrammetric point-clouds/meshes and extract object information, including individual tree locations and ground materials (Chen et al., 2019). The ultimate goal is to create realistic virtual environments and provide the necessary information for simulation. We tested the generalizability of the previously proposed framework using a database created under the U.S. Army's One World Terrain (OWT) project with a variety of landscapes (i.e., various buildings styles, types of vegetation, and urban density) and different data qualities (i.e., flight altitudes and overlap between images). Although the database is considerably larger than existing databases, it remains unknown whether deep-learning algorithms have truly achieved their full potential in terms of accuracy, as sizable data sets for training and validation are currently lacking. Obtaining large annotated 3D point-cloud databases is time-consuming and labor-intensive, not only from a data annotation perspective in which the data must be manually labeled by well-trained personnel, but also from a raw data collection and processing perspective. Furthermore, it is generally difficult for segmentation models to differentiate objects, such as buildings and tree masses, and these types of scenarios do not always exist in the collected data set. Thus, the objective of this study is to investigate using synthetic photogrammetric data to substitute real-world data in training deep-learning algorithms. We have investigated methods for generating synthetic UAV-based photogrammetric data to provide a sufficiently sized database for training a deep-learning algorithm with the ability to enlarge the data size for scenarios in which deep-learning models have difficulties.

## ABOUT THE AUTHORS

**Meida Chen** is currently a research associate at the University of Southern California's Institute for Creative Technologies (USC-ICT) working on the One World Terrain project. He received his Ph.D. degree at USC Sonny Astani Department of Civil and Environmental Engineering. His research focuses on the semantic modeling of outdoor scenes for the creation of virtual environments and simulations. Email: mechen@ict.usc.edu

**Andrew Feng** is currently a research scientist at USC-ICT working on the One World Terrain project. Previously, he was a research associate focusing on character animation and automatic 3D avatar generation. His research work involves applying machine learning techniques to solve computer graphics problems such as animation synthesis, mesh skinning, and mesh deformation. He received his Ph.D. and MS degree in computer science from the University of Illinois at Urbana-Champaign. Email: feng@ict.usc.edu

**Kyle McCullough** is the Director of Modeling & Simulation at USC ICT. His research involves geospatial initiatives in support of the Army's One World Terrain project, as well as advanced prototype systems development. His work includes utilizing AI and 3D visualization to increase fidelity and realism in large-scale dynamic simulation environments, and automating typically human-in-the-loop processes for Geo-specific 3D terrain data generation. He has published multiple papers on photogrammetric reconstruction, automated feature attribution, and autonomous agents. Kyle received awards from I/ITSEC and the Raindance festival, winning "Best Interactive Narrative VR Experience" in 2018. He has a B.F.A. from New York University.
Email: McCullough@ict.usc.edu

**Pratusha Bhuvana Prasad** is currently a researcher at USC-ICT working on the One World Terrain project. Her research focuses on computer vision for geometry and using machine learning methods to solve the same. She has a Master's degree from the Ming Hsieh Department of Electrical and Computer Engineering, USC. Email: bprasad@ict.usc.edu

**Ryan McAlinden** is a Senior Technology Advisor for the US Army's Synthetic Training Environment (STE), part of Army Futures Command (AFC). He is the Cross Functional Team (CFT) lead for the One World Terrain (OWT) initiative, which seeks to produce a high-resolution, geo-specific 3D representation of the surface used in the latest rendering engines, simulations and applications. Ryan previously served as Director of Modeling, Simulation & Training at the University of Southern California's Institute for Creative Technologies where he led several initiatives related to training modernization across the Services. Ryan rejoined ICT in 2013 after an assignment at the NATO Communications & Information Agency (NCIA) in The Hague, Netherlands. There he led the provision of operational analysis support to the International Security Assistance Force (ISAF) Headquarters in Kabul, Afghanistan. Ryan's research interests lie in the design, development, implementation and fielding of solutions that are at the cross-section of 3D rendering, geospatial science, and human-computer interaction. Ryan earned his B.S. from Rutgers University and M.S. in computer science from USC. Email: ryan.e.mcalinden.civ@mail.mil

**Lucio Soibelman** is a Professor and Chair of the Sonny Astani Department of Civil and Environmental Engineering at USC. Dr. Soibelman's research focuses on use of information technology for economic development, information technology support for construction management, process integration during the development of large-scale engineering systems, information logistics, artificial intelligence, data mining, knowledge discovery, image reasoning, text mining, machine learning, multi-reasoning mechanisms, sensors, sensor networks, and advanced infrastructure systems. Email: soibelman@usc.edu

# Generating Synthetic Photogrammetric Data for Training Deep Learning based 3D Point Cloud Segmentation Models

**Meida Chen, Andrew Feng, Kyle McCullough, Pratusha Bhuvana Prasad,**
***USC Institute for Creative Technologies***
**Los Angeles, California**
**{mechen, feng, McCullough, bprasad}@ict.usc.edu**

**Ryan McAlinden**
***Synthetic Training Environment Cross Functional Team***
**Los Angeles, California**
**ryan.e.mcalinden.civ@mail.mil**

**Lucio Soibelman**
***USC Department of Civil and Environmental Engineering***
**Los Angeles, California**
**soibelman@usc.edu**

## INTRODUCTION

Unmanned aerial vehicle (UAV)-based aerial image collection and photogrammetric techniques allow for rapid acquisition and reconstruction of high-fidelity and geo-specific 3D terrain data. These 3D data types provide the foundation for creating high-quality virtual environments for training and simulations (e.g., mission planning and rehearsal). Although the raw 3D photogrammetric data allows simple analytics in the generated virtual environments such as measuring distance or computing traversal time, they cannot enable sophisticated simulation of a realistic virtual battlespace such as assessing potential damages in the terrain or present structures. This is due to the fact that photogrammetric techniques cannot produce 3D data with semantic information (Chen, McAlinden, Spicer, & Soibelman, 2019; Chen, Feng, McAlinden, & Soibelman, 2020). At I/ITSEC 2019, we presented a fully-automated workflow to segment photogrammetric data using state-of-the-art deep-learning algorithms (Chen et al., 2019). The generalizability of the designed workflow was validated using a real-world UAV-based photogrammetric database that was created under the U.S. Army's One World Terrain (OWT) project. USC-ICT developed a research prototype named Semantic Terrain Point Labeling System Plus (STPLS+), which allows non-expert users to produce an attributed 3D model that is ready for simulation engine and intelligence analysis.

Despite the great success achieved, however, it is well known that deep-learning algorithms are data-hungry, especially in the 3D domain. Thus, data annotation plays a crucial role in training an effective deep-learning model for accurate prediction at run-time. Acquiring and annotating real-world 3D data is a labor-intensive process that requires many staff-hours and corresponding supervision to ensure that the resulting annotations are accurate. It is worth pointing out here that the USC-ICT research team have devoted considerable efforts over the past two years on establishing and annotating the real-world UAV-based photogrammetric database that was previously presented (Chen et al., Forthcoming; Chen et al., 2019). Several online crowdsourcing platforms exist, such as Amazon Mechanical Turk (MTurk), that were designed to aid in recruiting people to complete various annotation tasks. However, since certain sensitive data is restricted to official personnel, crowdsourcing is not a feasible solution in many cases. Furthermore, no effective/automatic approach to evaluate labeler performance exists. Thus, the evaluation process heavily relies on human judgment, which is labor-intensive and can be inaccurate in some cases.

To this end, this study designed and developed a synthetic 3D data generation framework. The framework was designed to exploit the synthetic photogrammetric data to train a deep-learning-based point-cloud segmentation model without the efforts for real-world collections and manual data annotations. To create the annotated 3D terrain, synthetic images are rendered based on simulated drone paths over a virtual environment. The rendered images are then used to produce a synthetic point-cloud with similar fidelity and quality as real-world UAV-based photogrammetric data. Ground-truth annotation is automatically obtained via a ray-casting and nearest neighbor search process.

Experiments were conducted to validate the designed framework and answer a set of fundamental questions related to how synthetic data should be generated. It is worth noting that regardless of the training data sources (i.e., whether it is synthetic training data or real-world training data), the performance of a trained point-cloud segmentation model has to be tested on the real-world dataset since the ultimate goal is to deploy the model in the real world. Thus, our previously created real-world UAV-based photogrammetric database is utilized again in this study. The experimental results showed that a model trained with the generated synthetic data was able to produce accurate segmentation on real-world UAV captured point-clouds.

## THE FRAMEWORK FOR GENERATING ANNOTATED SYNTHETIC PHOTOGRAMMETRIC DATA

The designed synthetic data-generation framework is illustrated in Figure 1, which emphasizes the main elements and steps involved in the process. The designed framework can be used to generate synthetic training data in which labels can be automatically created during the data generation process. Please note that this research is not intended to generate synthetic data with realistic appearances to human beings. However, the generated data should have similar enough features such that deep-learning models trained using the synthetic data can achieve a similar performance to that of real-world data.

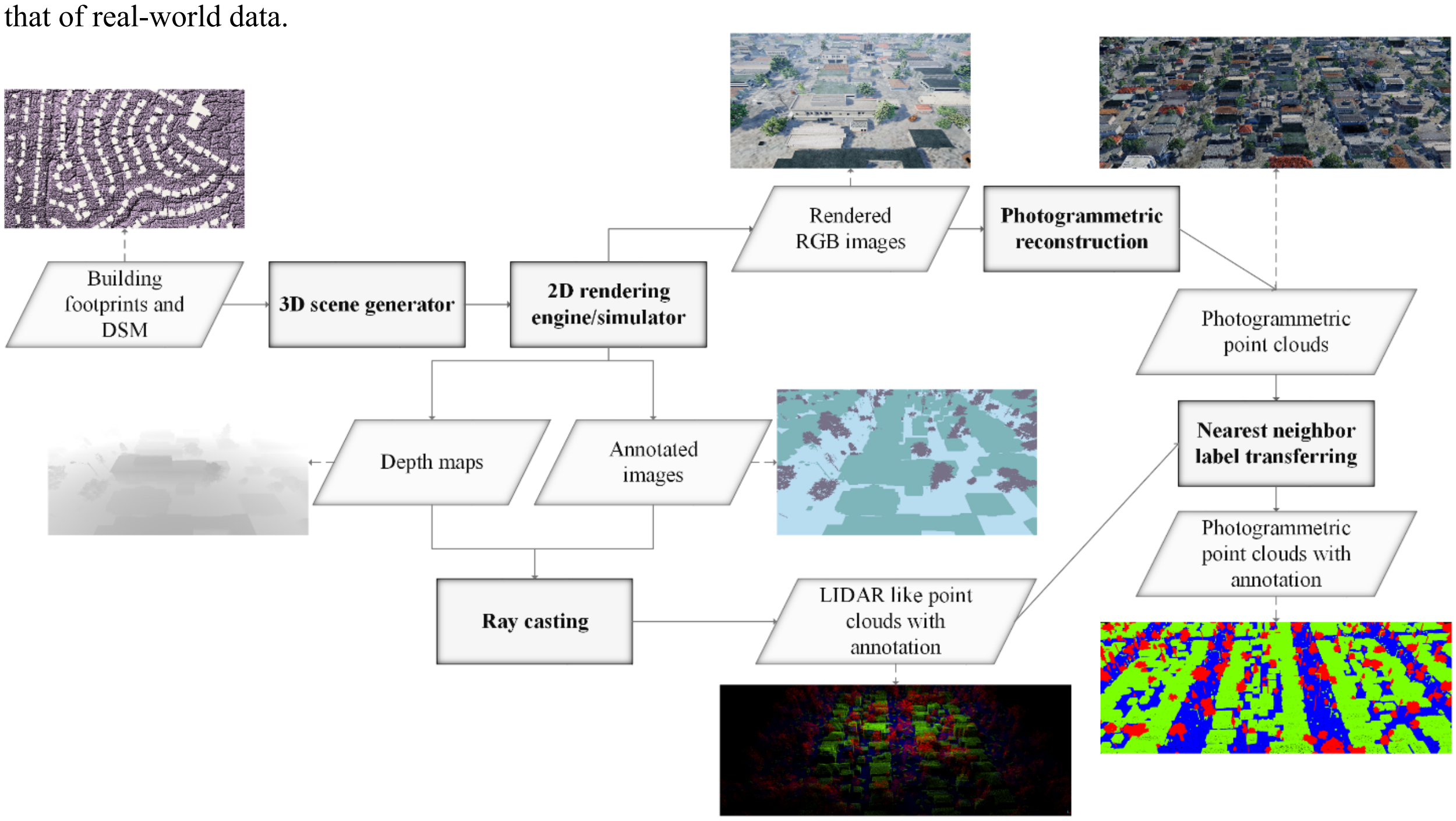

**Figure 1. The designed synthetic data generation framework**

First, 3D scenes were generated. Digital surface models (DSM) were gathered from publicly available GIS data sources. Buildings models were procedurally generated with predefined rules and open-sourced building footprints as inputs. To complete the 3D scene, artists created geotypical 3D models of clutter and small objects (e.g., trees, light poles, parking meters, street signs, cars, and so forth) that were procedurally placed in the scene. Next, the generated 3D terrain models were imported into game engines for the rendering process. AirSim (i.e., an open-source simulator for autonomous vehicles) was used to render images and generate associated ground-truth labels. The rendered images were further processed using commercial photogrammetry software Bentley ContextCapture (Bentley, 2020). Finally, labels from 2D images were projected to a 3D point-cloud using a ray casting method and the nearest neighbor algorithm.

### The 3D Scene Generation Process

Procedurally generating 3D scenes with the desired randomness is the key element in the designed framework. The main objects that need placement in the scene include the terrain surface model, buildings, vegetation, vehicles, and city clutter. Since we intended to design the framework to generate a large database of 3D scenes, scalability played a key role in the design process. Thus, user intervention during the data generation process was minimized with all required input data easily obtained. DSM can be obtained from the National Elevation Dataset (NED), which is the terrain surface elevation data produced and distributed by the USGS (Gesch, Evans, Mauck, Hutchinson, & Carswell Jr, 2009). It is worth pointing out that the highest resolution data in the NED is 1 meter. Although one-meter DSM resolution can already satisfy many different applications, it is still considered low when creating a synthetic 3D scene for image rendering and 3D reconstructions. Thus, the DSM first needs to be up-sampled. Fine details were then added to the terrain by raising or lowering the terrain elevation to create 3D geometry that is similar to ditches, street gutters, deceleration strips, and so forth. Figure 2 shows an example of a DSM gathered from the NED and the modified DSM.

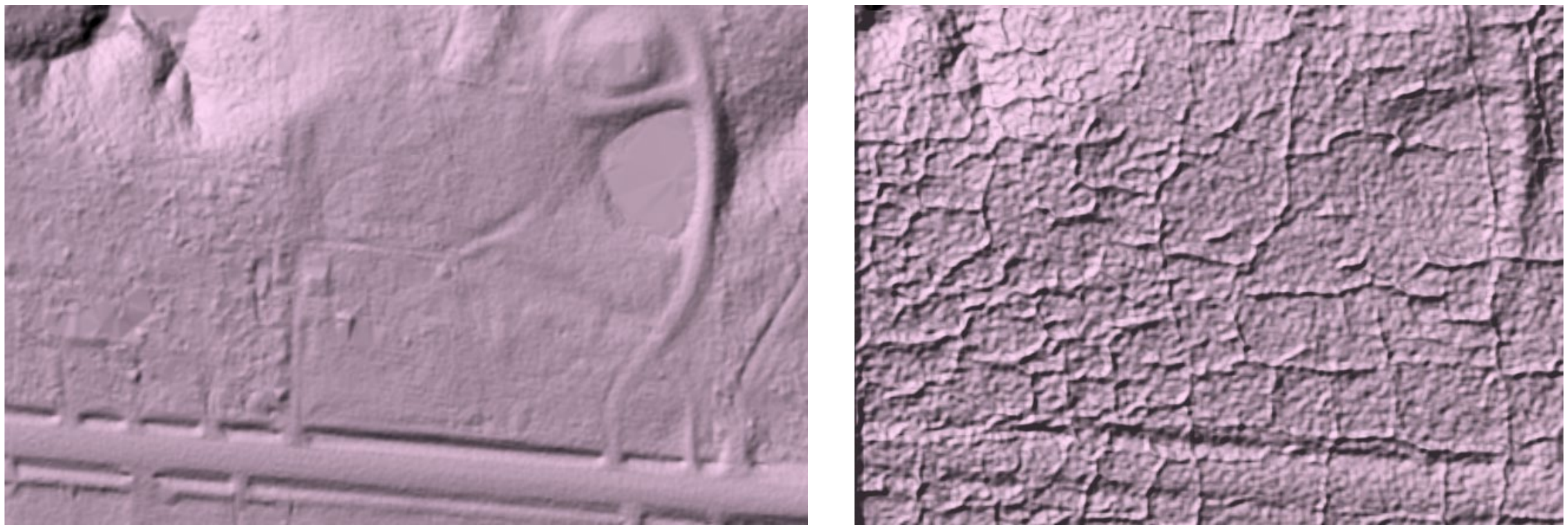

(a) (b)

**Figure 2. DSM: (a) the original DSM from the NED, and (b) modified DSM**

To create 3D building models, we used commercial software GameSim's Procedural Model Generation (GameSim, 2020)–in this study. The 3D building models were generated based on the given building footprints which were gathered from Open Street Map (OSM). Rather than using original preprogrammed OSM heights, building heights were randomly assigned to each building footprint to ensure randomness. Footprints were extruded to represent the basic 3D building models, followed by procedurally adding windows, doors, and other architectural elements to the building façades using a set of parametric rules. As a simple example of how this works, one can imagine a building façade represented by a rectangle, with the floors and tiles of windows and doors split along vertical and horizontal directions by predefined values. Roofs were randomly selected for each building from predefined roof styles (i.e., flat roof, flat roof with parapet, gable roof, and hip roof), and roof elements (i.e., chimneys, exhaust vents, and so forth) and architectural features (i.e., dormers, cones, turrets, and so forth) on top of the roof. Figure 3 shows examples of 3D models that are procedurally generated with the same building footprint.

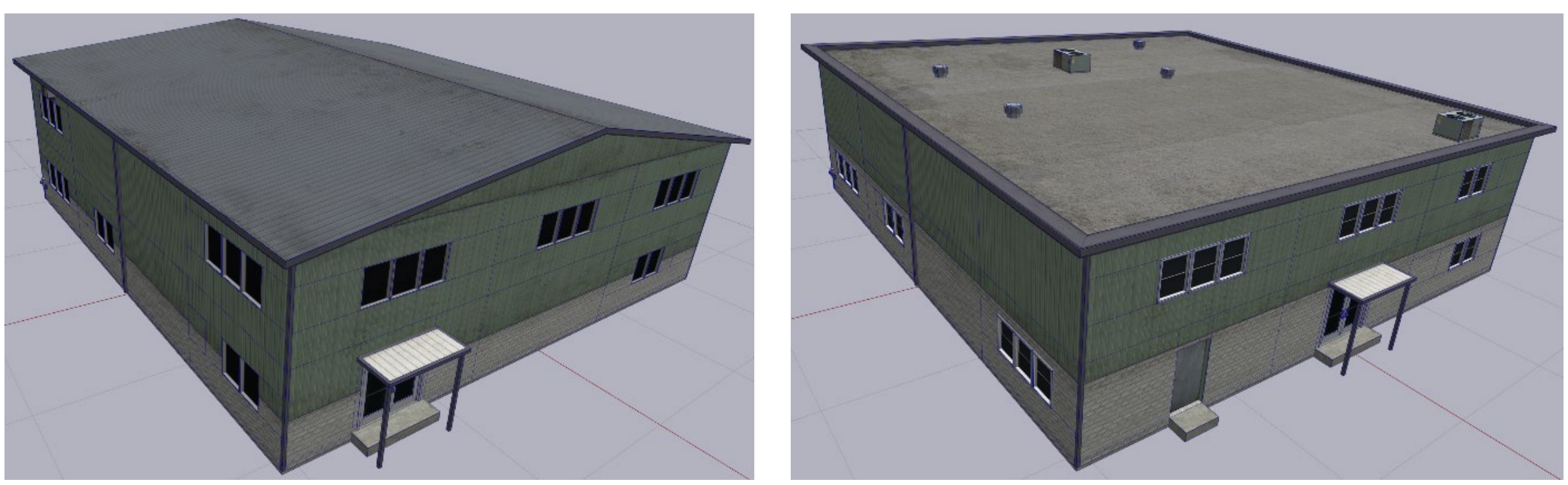

(a) (b)

**Figure 3. Procedurally generated 3D building models with the same building footprint**

Next, we placed vegetation, vehicles, and city clutter within the scene in randomly generated positions with randomly selected scale values within predefined ranges. To ensure that objects did not intersect (e.g., cars not physically intersecting with trees) a minimum distance constraint was enabled while generating object positions. In addition, any generated object positions inside the building footprints were removed. To create forests instead of individual trees, polygons were randomly created as boundaries of forest and dense tree positions were generated within the boundary.

**2D Image Rendering and 3D Point-cloud Reconstruction**

To render 2D images for the photogrammetric reconstruction of 3D terrains, the generated 3D building models and DSM must be imported into a game engine, and 3D objects (e.g., trees and vehicles) have to be placed at the generated positions. In this study, we used 3D models that were easily obtained from the 3D model marketplace. To ensure the realness of the generated scene, both gravity and physical collisions were enabled so that vehicles were correctly orientated to follow the terrain slope. Figure 4 shows the generated 3D scenes using the created object positions and scales.

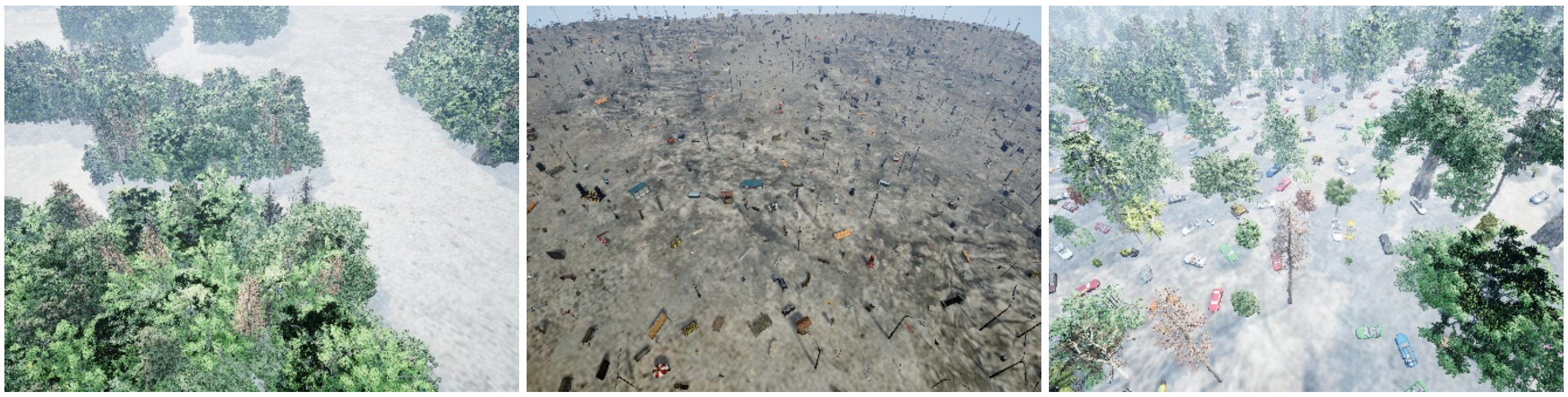

(a) (b) (c)

**Figure 4. Generated 3D scenes: (a) forests (b) city clutter, and (c) trees and vehicles**

The high-fidelity visual and physical simulator—i.e., AirSim (Shah, Dey, Lovett, & Kapoor, 2018)—was utilized for 2D image-rendering purposes. AirSim was originally designed for developing and testing autonomous driving vehicle algorithms. Since then, it has been widely used for solving other computer vision problems such as semantic segmentation of 2D images (Zhang, David, & Gong, 2017), real-time monocular depth estimation (Atapour-Abarghouei & Breckon, 2018), and aerial path planning optimization (Smith, Moehrle, Goesele, & Heidrich, 2019). The three main outputs of AirSim include photorealistic 2D images and their associated annotation and depth map. Figure 5 shows the outputs of AirSim with the generated 3D scene imported. The same UAV path-planning algorithm used in real-world UAV captures was used for rendering 2D images in the virtual environment. We also used crosshatch UAV paths in this study. Both front and side overlap between images was used during the rendering process to ensure the 3D reconstruction quality.

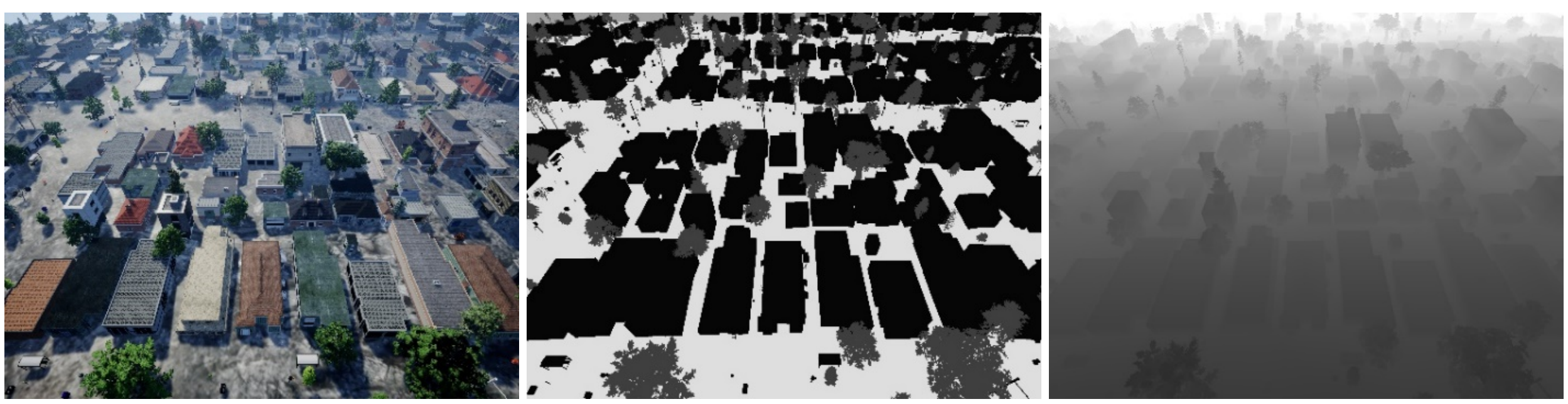

(a) (b) (c)

**Figure 5. Outputs from the simulator (i.e., AirSim): (a) rendered image; (b) annotation; and (c) depth map**

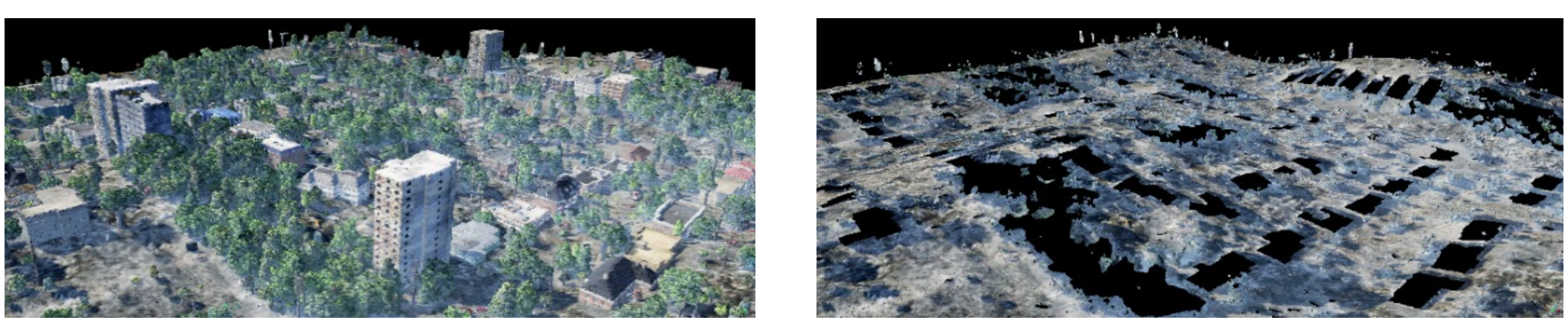

(a) (b)

**Figure 6. Photogrammetric point-cloud annotation using ray casting: (a) raw photogrammetric point-cloud and (b) extracted ground points**

We used commercial photogrammetry software Bentley ContextCapture for the photogrammetric reconstruction process. Annotating the generated point-clouds is an important task. Other researchers have pointed out that, noises were introduced during the photogrammetric-reconstruction process due to the generation of mismatches (Rupnik, Nex, & Remondino, 2014). Consequently, the 3D geometry of the reconstructed point-cloud was not accurate, and simply utilizing a ray casting approach to project the 2D ground truth labels to the photogrammetric generated 3D point-clouds did not provide accurate annotation. Figure 6 illustrates such an issue in which the ground points are labeled and extracted by projecting 2D labels through the ray casting method with intrinsic and extrinsic camera parameters produced from the bundle adjustment process. The extracted ground points have noises floating on air as shown in Figure 6 (b).

To overcome such a challenge, a ground truth 3D point-cloud was first created using the ray casting method with depth maps generated directly from the simulator. Following that, a k-nearest neighbor algorithm was used to transfer the point labels from the depth map-created point-cloud to the photogrammetric point-cloud. Figure 7 shows the annotated point-clouds created from depth maps and extracted ground points from the annotated point-cloud using the k-nearest neighbor algorithm. Comparing the results in Figure 6 (b) and Figure 7 (b), annotation noises can be eliminated by using the proposed k-nearest neighbor approach.

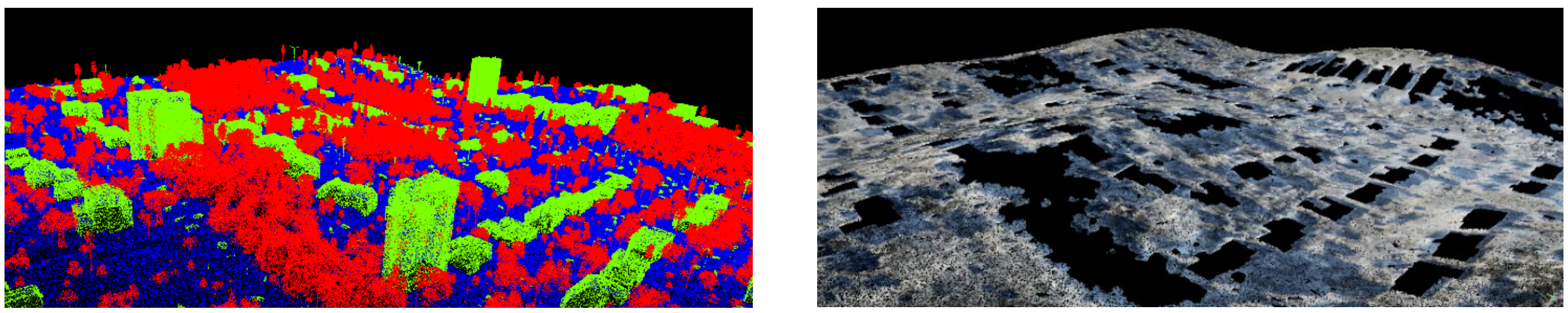

(a) (b)

**Figure 7. Photogrammetric point-cloud annotation using a k-nearest neighbor algorithm: (a) depth map generated point-cloud with annotation and (b) extracted ground points from an annotated photogrammetric point-cloud**

**EXPERIMENTS AND RESULTS**

In this section, we provide an evaluation of the proposed method for generating annotated data for training deep-learning algorithms to segment 3D photogrammetric point-clouds. We conducted the experiments to answer a set of fundamental questions related to how synthetic data should be generated. We used the deep-learning-based point-cloud segmentation algorithm (i.e., 3D U-net) that was introduced on previous work (Chen et al., 2019) in this study in order to test the synthetic training data.

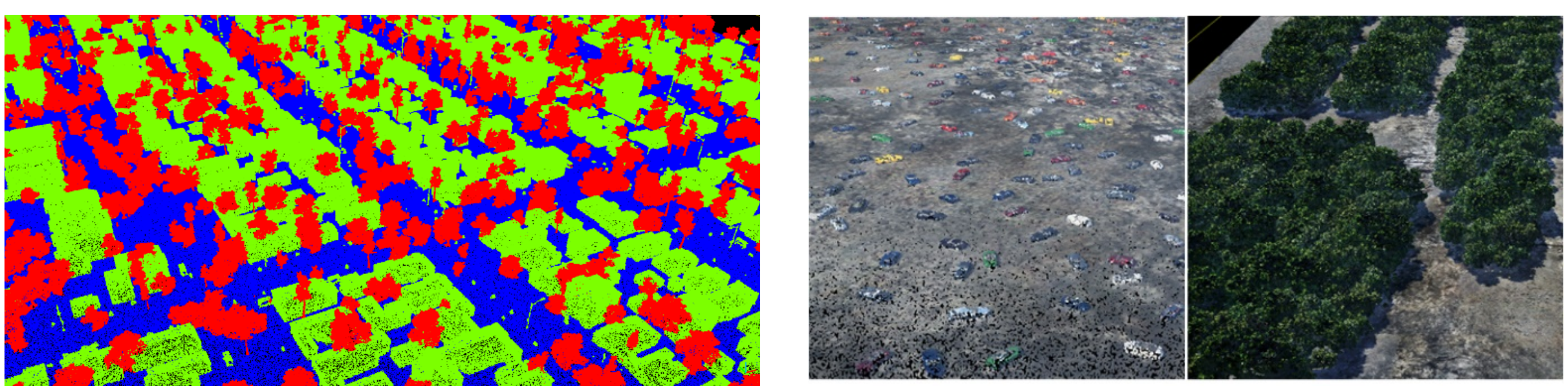

(a) (b)

**Figure 8. Generated synthetic training data sets: (a) The first synthetic training data set; (b) forests and vehicles in the second synthetic training data set**

Four sets of synthetic training data were generated for this study. In the first synthetic training data set, the DSM was obtained from the NED without any modifications (i.e., DSM with smooth surfaces and 1-meter resolution). 3D building models were created with OSM footprints and basic parametric rules for adding windows and doors. The generated 3D building models did not contain any complex architectural elements (e.g., protruding balcony). To create realistic contextual relationships between objects, city clutter (e.g., street signs, traffic lights, light poles, bus stops, and so forth) was placed in the scene along the road vectors obtained from OSM, and individual trees were placed around the buildings. Figure 8 (a) shows an example of the generated scene in the first synthetic training data set. In the second synthetic training data set, the DSM was modified to add fine details as discussed in "the 3D scene generation process" section. Forests and vehicles were added as separate scenes to the training data sets to increase the scene complexity. Figure 8 (b) shows the added forest and vehicle scenes.

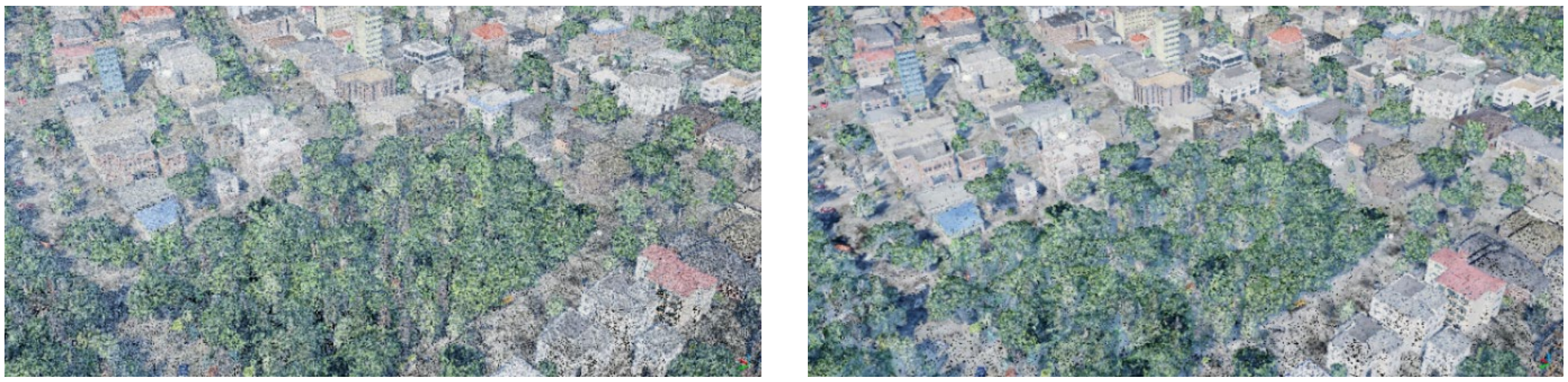

(a) (b)

**Figure 9. Generated synthetic training data sets: (a) the third synthetic training data set; and (b) the fourth synthetic training data set**

In the third and fourth synthetic training data sets, the 3D scenes were created following the procedures introduced in "the 3D Scene Generation Process" and "2D Image Rendering and 3D Point-cloud Reconstruction" sections. Complex architectural elements were added while creating 3D building models. Unlike the first and second training data sets in which city clutter was placed along the roads, in the third and fourth data sets, we randomly placed the clutter, individual trees, and vehicles in the scenes so that they did not have realistic contextual relationships between objects. Differences between the third and fourth data sets included creating the point-cloud directly from the rendered depth maps in the third data set, while the point-clouds in the fourth data set were created through the photogrammetric reconstruction and the k-nearest neighbor labeling processes. Figure 9 (a) and (b) show example scenes in the third and fourth synthetic training data sets.

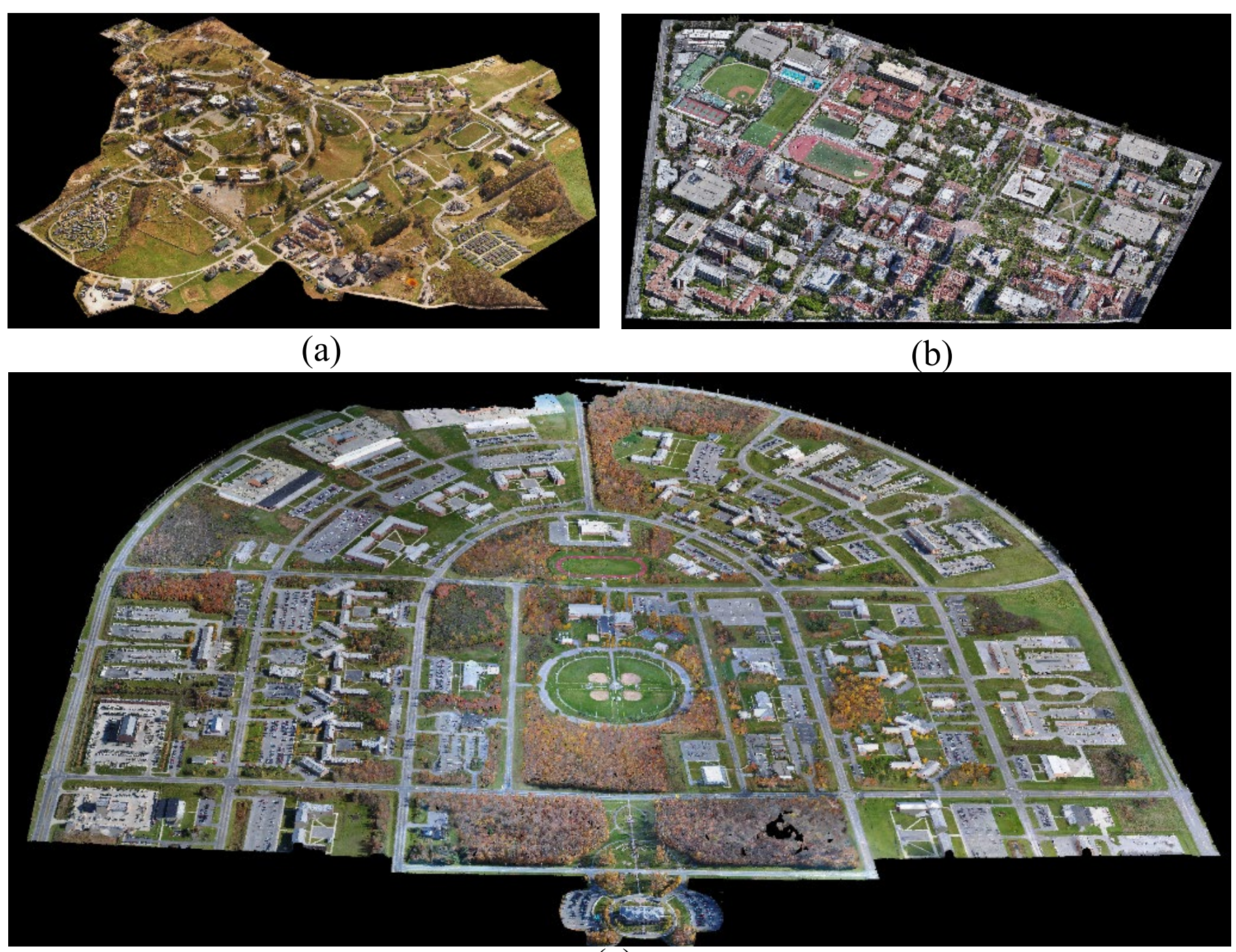

(a) (b) (c)

**Figure 10. Point-clouds of real-world data sets for testing: (a) Fort Drum; (b) MUTC; (c) USC**

Fort Drum, MUTC, and USC were selected from the real-world UAV-based photogrammetric database as the testing cases. Figure 10 (a), (b), and (c) show the point-clouds of Fort Drum, MUTC, and USC, respectively. We trained five 3D U-net models using the four generated synthetic training data sets and the real-world training data set with the same hyperparameter values. The models were then applied on Fort Drum, MUTC, and USC data sets to compare performances. The commonly used precision, recall, harmonic mean of precision and recall (i.e., F1 score), and intersection over union (IOU), also known as the Jaccard Index, were used to evaluate the segmentation results. IOU is computed as the points overlapped between the predicted segmentation and the ground truth divided by the points union between the predicted segmentation and the ground truth. Segmentation results are summarized in Tables 1-5.

**Table 1. Segmentation results with the first synthetic training data set**

| | precision | recall | f1-score | IOU |
|---|---|---|---|---|
| **Fort Drum** | | | | |
| ground | 0.944 | 0.863 | 0.901 | 0.82 |
| building | 0.409 | 0.861 | 0.555 | 0.384 |
| tree | 0.953 | 0.589 | 0.728 | 0.572 |
| macro avg | 0.769 | 0.771 | 0.728 | 0.592 |
| weighted avg | 0.868 | 0.781 | 0.799 | 0.683 |
| **MTUC** | | | | |
| ground | 0.969 | 0.794 | 0.873 | 0.774 |
| building | 0.469 | 0.867 | 0.609 | 0.438 |
| tree | 0.876 | 0.728 | 0.795 | 0.66 |
| macro avg | 0.771 | 0.796 | 0.759 | 0.624 |
| weighted avg | 0.869 | 0.798 | 0.816 | 0.7 |
| **USC** | | | | |
| ground | 0.877 | 0.871 | 0.874 | 0.776 |
| building | 0.888 | 0.844 | 0.865 | 0.763 |
| tree | 0.809 | 0.88 | 0.843 | 0.728 |
| macro avg | 0.858 | 0.865 | 0.861 | 0.756 |
| weighted avg | 0.863 | 0.861 | 0.861 | 0.756 |

**Table 2. Segmentation results with the second synthetic training data set**

| precision | recall | f1-score | IOU |
|---|---|---|---|
| **Fort Drum** | | | |
| 0.942 | 0.97 | 0.956 | 0.915 |
| 0.808 | 0.795 | 0.802 | 0.669 |
| 0.933 | 0.888 | 0.91 | 0.835 |
| 0.894 | 0.884 | 0.889 | 0.806 |
| 0.92 | 0.92 | 0.92 | 0.855 |
| **MTUC** | | | |
| 0.967 | 0.961 | 0.964 | 0.93 |
| 0.863 | 0.793 | 0.826 | 0.704 |
| 0.814 | 0.923 | 0.865 | 0.763 |
| 0.881 | 0.892 | 0.885 | 0.799 |
| 0.928 | 0.926 | 0.926 | 0.868 |
| **USC** | | | |
| 0.886 | 0.94 | 0.912 | 0.839 |
| 0.938 | 0.746 | 0.831 | 0.711 |
| 0.712 | 0.915 | 0.801 | 0.668 |
| 0.845 | 0.867 | 0.848 | 0.739 |
| 0.861 | 0.842 | 0.843 | 0.73 |

**Table 3. Segmentation results with the third synthetic training data set.**

| | precision | recall | f1-score | IOU |
|---|---|---|---|---|
| Fort Drum | | | | |
| ground | 0.871 | 0.975 | 0.92 | 0.852 |
| building | 0.351 | 0.872 | 0.501 | 0.334 |
| tree | 0.984 | 0.049 | 0.093 | 0.049 |
| macro avg | 0.736 | 0.632 | 0.505 | 0.412 |
| weighted avg | 0.829 | 0.684 | 0.613 | 0.537 |
| MTUC | | | | |
| ground | 0.953 | 0.976 | 0.965 | 0.932 |
| building | 0.551 | 0.895 | 0.682 | 0.518 |
| tree | 0.967 | 0.08 | 0.148 | 0.08 |
| macro avg | 0.824 | 0.65 | 0.598 | 0.51 |
| weighted avg | 0.885 | 0.84 | 0.804 | 0.744 |
| USC | | | | |
| ground | 0.866 | 0.932 | 0.898 | 0.814 |
| building | 0.747 | 0.949 | 0.836 | 0.718 |
| tree | 0.978 | 0.474 | 0.639 | 0.469 |
| macro avg | 0.863 | 0.785 | 0.791 | 0.667 |
| weighted avg | 0.842 | 0.81 | 0.795 | 0.671 |

**Table 4. Segmentation results with the fourth synthetic training data set.**

| precision | recall | f1-score | IOU |
|---|---|---|---|
| Fort Drum | | | |
| 0.926 | 0.98 | 0.952 | 0.908 |
| 0.837 | 0.824 | 0.83 | 0.71 |
| 0.982 | 0.884 | 0.93 | 0.87 |
| 0.915 | 0.896 | 0.904 | 0.829 |
| 0.93 | 0.928 | 0.928 | 0.868 |
| MTUC | | | |
| 0.956 | 0.972 | 0.964 | 0.93 |
| 0.863 | 0.845 | 0.854 | 0.745 |
| 0.953 | 0.895 | 0.923 | 0.858 |
| 0.924 | 0.904 | 0.914 | 0.844 |
| 0.939 | 0.94 | 0.939 | 0.888 |
| USC | | | |
| 0.854 | 0.963 | 0.905 | 0.827 |
| 0.936 | 0.902 | 0.919 | 0.85 |
| 0.926 | 0.879 | 0.902 | 0.821 |
| 0.905 | 0.915 | 0.909 | 0.833 |
| 0.913 | 0.911 | 0.911 | 0.836 |

**Table 5. Segmentation results with the real-world training data set.**

| precision | recall | f1-score | IOU |
|---|---|---|---|
| Fort Drum | | | |
| 0.928 | 0.978 | 0.952 | 0.908 |
| 0.786 | 0.833 | 0.809 | 0.679 |
| 0.967 | 0.84 | 0.899 | 0.817 |
| 0.893 | 0.884 | 0.887 | 0.801 |
| 0.918 | 0.916 | 0.915 | 0.848 |
| MTUC | | | |
| 0.965 | 0.974 | 0.969 | 0.94 |
| 0.878 | 0.878 | 0.878 | 0.782 |
| 0.933 | 0.888 | 0.91 | 0.834 |
| 0.925 | 0.913 | 0.919 | 0.852 |
| 0.945 | 0.945 | 0.945 | 0.898 |
| USC | | | |
| 0.866 | 0.944 | 0.903 | 0.823 |
| 0.941 | 0.869 | 0.904 | 0.824 |
| 0.869 | 0.909 | 0.888 | 0.799 |
| 0.892 | 0.907 | 0.898 | 0.815 |
| 0.902 | 0.899 | 0.899 | 0.817 |

**Question 1. How Much Does It Help to Add Details to the Synthetic Scene?**

Tables 1 and 2 show a direct comparison between adding and not adding fine details to the DSM of the synthetic training data. Ground segmentation performance improved for all three testing data sets (i.e., F1-score improves from 0.901 to 0.956 in the Fort Drum data set, from 0.873 to 0.964 in the MUTC data set and from 0.874 to 0.912 in the USC data set). In addition, since the forest scene was added to the second synthetic training data set, the model trained with the second set was more robust for segmenting forests from buildings. As such, the performance of segmenting both forests and buildings improved, which performance improvements are shown in the Fort Drum and the MUTC segmentation results. The F1-score of building segmentation improved from 0.555 to 0.802 in the Fort Drum data set and from 0.609 to 0.826 in the MUTC data set. The F1-score of vegetation segmentation improved from 0.728 to 0.910 in the Fort Drum data set and from 0.795 to 0.865 in the MUTC data set. However, in the USC data set, the performance of segmenting buildings and trees decreased due to the following two reasons. First, no forests exist on the USC campus, and most of the trees are planted at predetermined intervals along streets. Second, buildings on the USC campus include more complex architectural elements (e.g., protruding balconies and church towers) than the generated synthetic buildings, which caused confusion in the model trained on the synthetic data between complex buildings and forests. Therefore, adding details to synthetic buildings is also a necessary step. The segmentation performance using synthetic buildings with complex architectural elements will be discussed later in this section in the answer to Question 3.

**Question 2. Is It Necessary to Use Photogrammetric Reconstructed Point-clouds Instead of the Depth Map-generated Point-clouds for Training Purposes?**

From the proposed synthetic data-generation framework shown in Figure 1, readers can see that annotated 3D point-clouds can be directly generated from the rendered depth maps and 2D annotations. Figure 9 (a) and (b) shows that the point-clouds generated from depth maps and photogrammetry reconstruction process have almost no visual differences. These concerns prompted us to examine the performance differences between training a model with the depth map-generated point-clouds and the photogrammetric reconstructed point-clouds. Tables 3 and 4 show a direct comparison between the segmentation results using the two types of training data for the models. Overall, we can clearly see the benefits of using the photogrammetric-reconstructed point-clouds as the training data. The macro average of the F1-score improved from 0.505 to 0.904 in the Fort Drum data set, from 0.598 to 0.914 in the MUTC data set, and from 0.791 to 0.909 in the USC data set. Note that the vegetation IOU in all three data sets is very low (i.e., 0.049 in Fort Drum, 0.080 in MUTC and 0.469 in USC), because the quality of the depth map-generated point-clouds is different from the quality of the photogrammetry-reconstructed point-clouds. Figure 11 illustrates the tree point-clouds generated using depth maps and photogrammetry reconstruction. The photogrammetry reconstructed tree point-cloud appears as a solid blob with no points generated inside the crown. The depth map-generated tree point-cloud has a similar quality as LIDAR-collected data in which data points on the leaves inside of the crown can also be generated. Consequently, the segmentation model trained on depth map-generated point-clouds learned to predict points that were not on hollow-shaped objects as tree points. Therefore, rendering 2D RGB images and creating the training point-clouds from photogrammetry reconstruction is a necessary step to introduce photogrammetric noises to the data and improve the segmentation performance.

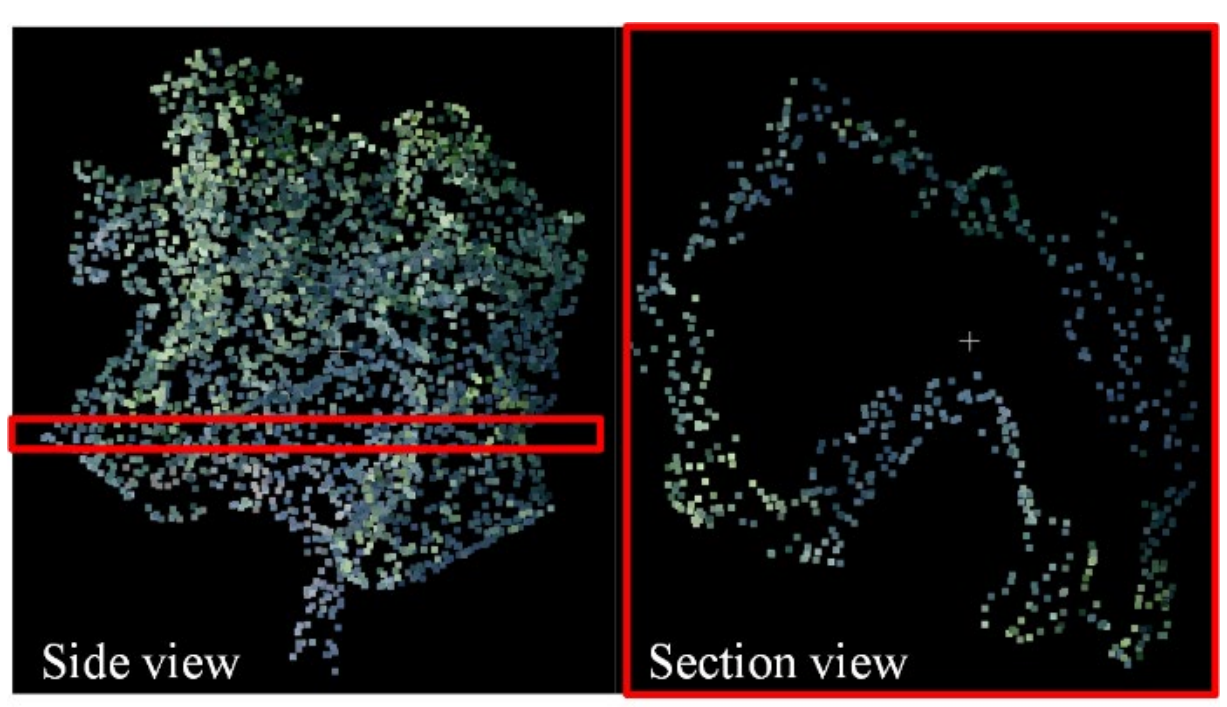

(a)

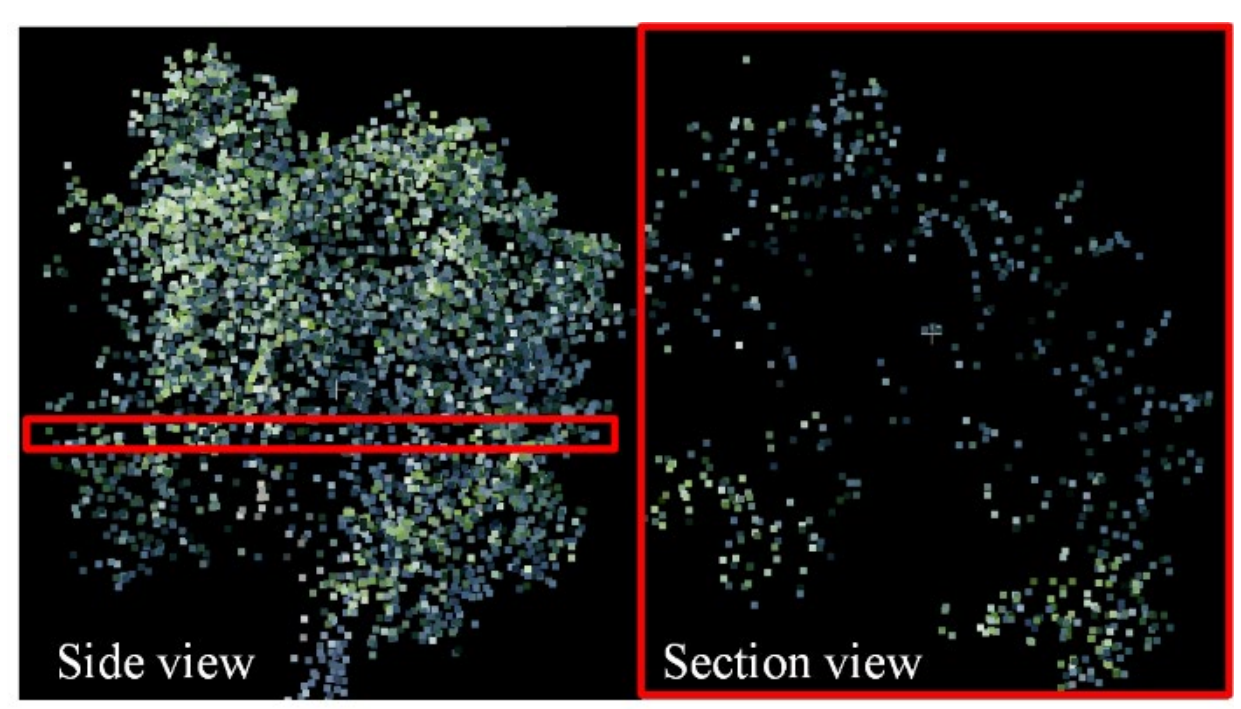

(b)

**Figure 11. Tree point-cloud: (a) depth map-generated tree point-cloud, and (b) photogrammetric-reconstructed tree point-cloud**

**Question 3. Is It Necessary to Create Synthetic Scenes with Realistic Contextual Relationships Between Objects?**

An interesting question is whether the generated synthetic scene should contain contextual relationships between objects similar to the real world. On the one hand, one could argue that it makes more sense to feed the synthetic data with contextual relationships between objects so the trained model is well adapted for real-world data predictions (e.g., the model understands that traffic lights usually appear near a cross-section). On the other hand, it might make more sense to randomize the synthetic scene as much as possible so that the trained model is more robust for predicting point labels in a new environment. To this end, we compared the segmentation results using the fourth synthetic training data sets (i.e., city clutter placed randomly) shown in Table 4 with the segmentation results using the second synthetic training data sets (i.e., city clutter was placed along the street) shown in Table 2. The performance of the trained model using the fourth synthetic training data sets outperformed the other model in all three testing cases. The F1-score of building segmentation improved from 0.802 to 0.830 in the Fort Drum data set and from 0.826 to 0.854 in the MUTC data set. Therefore, adding contextual relationships between objects into the synthetic scenes did not improve the model performance. In contrast, placing the city clutter randomly can increase the geometric complexity of the scene and thus improve the model performance. Note that a large improvement in the building F1-score is shown in the USC case (i.e., an improvement from 0.831 to 0.919). This is because complex architectural elements were also added into the fourth synthetic data sets, and the trained model correctly predicted buildings with complex architectural elements on the USC campus.

### Question 4. Can Synthetic Data Be Used for Training Deep-learning Models and Replace the Need for Creating Real-World Training Data?

Finally, Tables 4 and 5 provide a direct comparison between using synthetic training data and real-world training data for training point-cloud segmentation models. Note that the synthetic training data was created to cover a similar area size as the real-world training data to ensure an equivalent comparison. The models trained with the synthetic training data outperformed the models trained using the real-world training data in two out of three testing cases. Using the synthetic training data, the F1-score macro average improved from 0.887 to 0.904 in the Fort Drum case and from 0.898 to 0.909 in the USC case. Note that the model trained using the synthetic training data underperformed compared with the model trained using the real-world training data in the MUTC case. This is because the MUTC data set includes small bushes that the real-world training data also has but the synthetic training data does not. ***Overall, the comparison results validated the proposed synthetic data generation workflow can be used to create training data for deep-learning models and replace the need for creating real-world training data.***

## CONCLUSION

Deep-learning algorithms are data-hungry, especially in the 3D domain. Acquiring and annotating 3D data is a labor-intensive and time-consuming process. To this end, this study designed and developed a synthetic 3D data generation workflow and investigated the potential of using synthetic photogrammetric data to substitute real-world data for training deep-learning algorithms. The designed synthetic data- generation framework takes full advantage of the off-the-shelf 3D scene generation engine, autonomous vehicles and UAV simulator, and photogrammetry software. The key elements in the designed framework include randomness in the generated scenes and adding photogrammetric noises to the synthetic point-clouds. We validated the designed framework through a comparison of 3D U-nets trained on synthetic data and on real-world training data, and we analyzed the experiment results to answer four fundamental questions on how synthetic data should be generated. We found that adding detail to the ground surfaces and building facades while generating the synthetic scenes is a necessary step in boosting the model's performance. We also found that using synthetic data without producing a point-cloud through the photogrammetric-reconstruction process (i.e., eliminating the photogrammetric noises) decreases the model's performance dramatically, particularly in segmenting vegetation. Furthermore, the results show that adding realistic contextual relationships between objects to the synthetic data does not improve the model's performance. Finally, by comparing synthetic data and real-world data as the training data, the results show that the designed synthetic data-generation workflow can be used to create training data for training 3D U-net and achieve a similar performance to training data collected from the real-world. However, this study has several limitations. First, the generated synthetic data does not include small bushes, and the trained models cannot differentiate small bushes and city clutter correctly. Second, realistic surface materials were not generated in the synthetic scene, and point-cloud segmentation architecture such as 3DMV that takes advantage of 2D texture features was not tested in the experiments. Third, we conducted the experiment to test a deep-learning segmentation algorithm with volumetric representation only, and other state-of-the-art algorithms (e.g., PointNet, SPG) that operate on an unordered point set were not tested.

To conclude this paper, we outline promising future directions of research on utilizing the designed synthetic 3D data generation framework. First, investigating and finding the optimal settings is crucial for the UAV flight path to map an area of interest. Conduct experiments and test different UAV flight path settings that can affect the quality of the reconstructed point clouds is a necessary step. Unlike most of the existing studies where the experiments were conducted in the real world with limited testing cases due to the physical restrictions, we plan to conduct the experiments with the synthetic data in the virtual world and provide a comprehensive analysis of different data collection strategies. Second, for many tasks, data for specific objects cannot be easily obtained. For instance, there are limited amounts of military-related objects (e.g., tank, fighter aircraft, etc.) even in the OWT data repository, which contains data for several military bases. In our future works, we plan to expand the synthetic 3D database to include different military-related objects for different tasks. Third, since users can easily make changes to the generated synthetic scenes, the designed framework can also be used for developing and validating 3D change detection algorithms. Finally, annotating and creating 3D data to perform instance segmentation is even challenging than point-wise segmentation task since each object has to be manually annotated. In our future works, we also plan to expand the capability of the designed synthetic 3D data generation framework for creating instance labels in the 3D point clouds.

## ACKNOWLEDGMENTS

The authors would like to thank the two primary sponsors of this research: Army Futures Command (AFC) Synthetic Training Environment (STE), and the Office of Naval Research (ONR). We would also like to acknowledge the assistance provided by 3/7 Special Forces Group (SFG), Naval Special Warfare (NSW), the National Training Center (NTC), and the US Marine Corps. This work is supported by University Affiliated Research Center (UARC) award W911NF-14-D-0005. Statements and opinions expressed and content included do not necessarily reflect the position or the policy of the Government, and no official endorsement should be inferred.

## REFERENCES

Atapour-Abarghouei, A., & Breckon, T. P. (2018). Real-time monocular depth estimation using synthetic data with domain adaptation via image style transfer. Paper presented at the *Proceedings of the IEEE Conference on Computer Vision and Pattern Recognition,* 2800-2810.

Bentley. (2020). ContextCapture - 3D. Retrieved from https://www.bentley.com/en/products/brands/contextcapture

Chen, M., Feng, A., McAlinden, R., & Soibelman, L. (2020). Photogrammetric point cloud segmentation and object information extraction for creating virtual environments and simulations. *Journal of Management in Engineering, 36*(2), 04019046.

Chen, M., Feng, A., McCullough, K., Bhuvana Prasad, P., McAlinden, R., & Soibelman, L. (Forthcoming). 3D photogrammetry point cloud segmentation using a model ensembling framework. *Journal of Computing in Civil Engineering,* doi:10.1061/(ASCE)CP.1943-5487.0000929

Chen, M., McAlinden, R., Spicer, R., & Soibelman, L. (2019). Semantic modeling of outdoor scenes for the creation of virtual environments and simulations. Paper presented at the *Proceedings of the 52nd Hawaii International Conference on System Sciences,*

Chen, M., Feng, A., McCullough, K., Prasad, P. B., McAlinden, R., Soibelman, L., & Enloe, M. (2019). Fully automated photogrammetric data segmentation and object information extraction approach for creating simulation terrain. Paper presented at the *Interservice/Industry Training, Simulation, and Education Conference (I/ITSEC),*

GameSim. (2020). Procedural model generation. Retrieved from https://www.gamesim.com/proceduralmodeling/

Gesch, D., Evans, G., Mauck, J., Hutchinson, J., & Carswell Jr, W. J. (2009). The national map—Elevation. *US Geological Survey Fact Sheet, 3053*(4)

Rupnik, E., Nex, F., & Remondino, F. (2014). Oblique multi-camera systems-orientation and dense matching issues. *The International Archives of Photogrammetry, Remote Sensing and Spatial Information Sciences, 40*(3), 107.

Shah, S., Dey, D., Lovett, C., & Kapoor, A. (2018). Airsim: High-fidelity visual and physical simulation for autonomous vehicles. Paper presented at the *Field and Service Robotics,* 621-635.

Smith, N., Moehrle, N., Goesele, M., & Heidrich, W. (2019). Aerial path planning for urban scene reconstruction: A continuous optimization method and benchmark. *ACM Transactions on Graphics (TOG), 37*(6), 183.

Zhang, Y., David, P., & Gong, B. (2017). Curriculum domain adaptation for semantic segmentation of urban scenes. Paper presented at the *Proceedings of the IEEE International Conference on Computer Vision,* 2020-2030.